# Distributed Bayesian Learning with Stochastic Natural Gradient Expectation Propagation and the Posterior Server

**Leonard Hasenclever**                                    HASENCLEVER@STATS.OX.AC.UK
*Department of Statistics, University of Oxford, United Kingdom*

**Stefan Webb**                                           STEFANW@ROBOTS.OX.AC.UK
*Department of Engineering Sciences, University of Oxford, United Kingdom*

**Thibaut Lienart**                                       LIENART@STATS.OX.AC.UK
*Department of Statistics, University of Oxford, United Kingdom*

**Sebastian Vollmer**                                     VOLLMER@STATS.OX.AC.UK
*Department of Statistics, University of Oxford, United Kingdom*

**Balaji Lakshminarayanan**                               BALAJILN@GOOGLE.COM
*DeepMind*

**Charles Blundell**                                      CBLUNDELL@GOOGLE.COM
*DeepMind*

**Yee Whye Teh**                                          Y.W.TEH@STATS.OX.AC.UK
*Department of Statistics, University of Oxford, United Kingdom*

## Abstract

This paper makes two contributions to Bayesian machine learning algorithms. Firstly, we propose stochastic natural gradient expectation propagation (SNEP), a novel alternative to expectation propagation (EP), a popular variational inference algorithm. SNEP is a black box variational algorithm, in that it does not require any simplifying assumptions on the distribution of interest, beyond the existence of some Monte Carlo sampler for estimating the moments of the EP tilted distributions. Further, as opposed to EP which has no guarantee of convergence, SNEP can be shown to be convergent, even when using Monte Carlo moment estimates. Secondly, we propose a novel architecture for distributed Bayesian learning which we call the posterior server. The posterior server allows scalable and robust Bayesian learning in cases where a data set is stored in a distributed manner across a cluster, with each compute node containing a disjoint subset of data. An independent Monte Carlo sampler is run on each compute node, with direct access only to the local data subset, but which targets an approximation to the global posterior distribution given all data across the whole cluster. This is achieved by using a distributed asynchronous implementation of SNEP to pass messages across the cluster. We demonstrate SNEP and the posterior server on distributed Bayesian learning of logistic regression and neural networks.

## 1. Introduction

Algorithms and systems for enabling machine learning from large scale data sets are becoming increasingly important in the era of Big Data. This has driven many developments, including various forms of stochastic gradient descent, parallel and distributed learning systems, use of GPUs, sketching, random Fourier features, divide-and-conquer methods, as well as various approximation schemes. These large scale machine learning systems have in turn driven significant advances across many data-oriented sciences and technologies, ranging from the biological sciences, neuroscience, social sciences, signal processing, speech processing, natural language processing, computer vision etc.

In this paper we will consider methods for large scale *Bayesian* machine learning. As opposed to the more common empirical risk minimisation or maximum likelihood approaches, where learning is phrased as finding a set of parameters optimal with respect to a data set and to a loss or likelihood function, Bayesian machine learning rests upon probabilistic models which capture the dependencies among all observed and unobserved variables, and where learning is phrased as computing the posterior distribution over unobserved variables (including both latent variables and model parameters) given the observed data.

The Bayesian framework can more fully capture the uncertainty in learnt parameters and prevent overfitting. In principle this allows the use of more complex and larger scale models. However, Bayesian approaches are generally more computationally intensive than optimisation-based ones, and have to date not led to methods which are as scalable.

For complex models, exact computation of the posterior distribution is intractable and approximate schemes such as variational inference (VI) (Wainwright and Jordan, 2008), Markov chain Monte Carlo (MCMC) (Gilks et al., 1996) and sequential Monte Carlo (Doucet et al., 2001) are needed. Scalable methods in both traditions include: stochastic variational inference (Hoffman et al., 2013, Mnih and Gregor, 2014, Rezende et al., 2014) which apply minibatch stochastic gradient descent (Robbins and Monro, 1951) to optimise the variational objective function, stochastic gradient MCMC (Welling and Teh, 2011, Patterson and Teh, 2013, Ding et al., 2014, Teh et al., 2015, Leimkuhler, B. and Shang, X., 2016, Ma et al., 2015, Li et al., 2016) which uses minibatch stochastic gradients within MCMC, austerity MCMC (Korattikara et al., 2014, Bardenet et al., 2014) which uses data subsampling to reduce computational cost of Metropolis-Hastings acceptance steps, and embarrassingly parallel MCMC (Huang and Gelman, 2005, Scott et al., 2013, Wang and Dunson, 2013, Neiswanger et al., 2014) which distribute data across a cluster, runs independent MCMC samplers on each worker and combines samples across the cluster only at the end to reduce network communication costs. In addition, standard learning schemes have also been successfully deployed in large scale settings. A successful example that is particularly relevant to our work is expectation propagation (EP), introduced by (Minka, 2001) and (Opper and Winther, 2000) which iteratively constructs approximations to a factorized distribution. EP is at the heart of the TrueSkill XBox player rating and matching system (Herbrich et al., 2007).

Our work builds upon prior work on using EP for performing distributed Bayesian learning (Xu et al., 2014, Gelman et al., 2014). In this framework, a data set is partitioned into disjoint subsets with each subset stored on a worker node in a cluster. Learning is performed at each worker based on the data subset there using MCMC sampling. As opposed to embarrassingly parallel MCMC methods which only communicate the samples to the master at the end of learning, EP is used to communicate messages (infrequently) across the cluster. These messages coordinate the samplers such that the target distributions (which coincidentally are the tilted distributions in EP) of all samplers on all workers share certain moments, e.g. means and variances, hence the name sampling via moment sharing (SMS) coined by (Xu et al., 2014). At convergence, it can be shown that the target distributions of the samplers also share moments with the EP approximation to the global posterior distribution given all data, hence the target distributions on the workers can themselves be treated as approximations to the global posterior.

While SMS works well on simpler models like Bayesian logistic regression, we have found that it did not work for more complex, high-dimensional, and non-convex models like Bayesian deep neural networks. This is due to the non-convergence of EP, particularly as the moments of the tilted distributions needed by EP are estimated using MCMC sampling, with estimation noise that further compounds the well-known lack of convergence guarantees for EP, and the fact that extremely long MCMC runs are needed for the samplers to equilibrate due to the complex posterior distribution in these models.

Our first contribution is thus the development of stochastic natural-gradient EP (SNEP), an alternative algorithm to power EP (a generalisation of EP) (Minka, 2004) which optimises the same variational objective function. SNEP is a double-loop algorithm with convergence guarantees. The inner loop is a stochastic natural-gradient descent algorithm which tolerates estimation noise, so that SNEP is convergent even with moments estimated using MCMC samplers. Our derivation of SNEP improves upon the derivation of the convergent EP algorithm of (Heskes and Zoeter, 2002) in that ours works for a more general class of models, we make explicit the underlying variational objective function that is being optimised, and ours uses a natural-gradient descent algorithm (Amari and Nagaoka, 2001) more tolerant of Monte Carlo noise.

Building upon the development of SNEP, our second contribution is a distributed Bayesian learning architecture which we call the posterior server. In analogy to the parameter server (Ahmed et al., 2012) which maintains and serves the parameter vector to a cluster of workers, the posterior server maintains and serves (an approximation to) the posterior distribution. Figure 1 gives a schematic for the steps involved. Each worker has a subset of data, from which we get a likelihood function. It also maintains a tractable approximation of the likelihood and a cavity distribution which is effectively a conditional distribution over the parameters given all data on other workers. An MCMC sampler targets the normalised product of the cavity distribution and the (true) likelihood, and forms a stochastic estimate of the required moments, which is in turn used to update the likelihood approximation using stochastic natural-gradient descent. Each worker communicates with the posterior server asynchronously and in a non-blocking manner, sending the current likelihood approximation and receiving the new cavity distribution. This communication protocol makes more efficient use of computational resources on workers than SMS, which requires either synchronous or blocking asynchronous protocols.

Note that our set-up of the distributed learning problem is that each worker has access to a subset of data, and no worker has access to all data. This situation might occur in situations other than large scale learning. For example, when working with sensitive patient data which cannot be shared directly, we might still want to be able to make use of all available data across multiple sites to improve inference. Typically known as divide-and-conquer or consensus inference (Zhang et al., 2015b, Zhao et al., 2016, Kleiner et al., 2014, Battey et al., 2015), it is also well-known that this situation is harder than typical distributed learning settings where it is assumed that all data is accessible on all workers, which are settings assumed in DistBelief networks (Dean et al., 2012) and elastic-averaging SGD (Zhang et al., 2015a), two state-of-the-art distributed learning algorithms for neural networks.

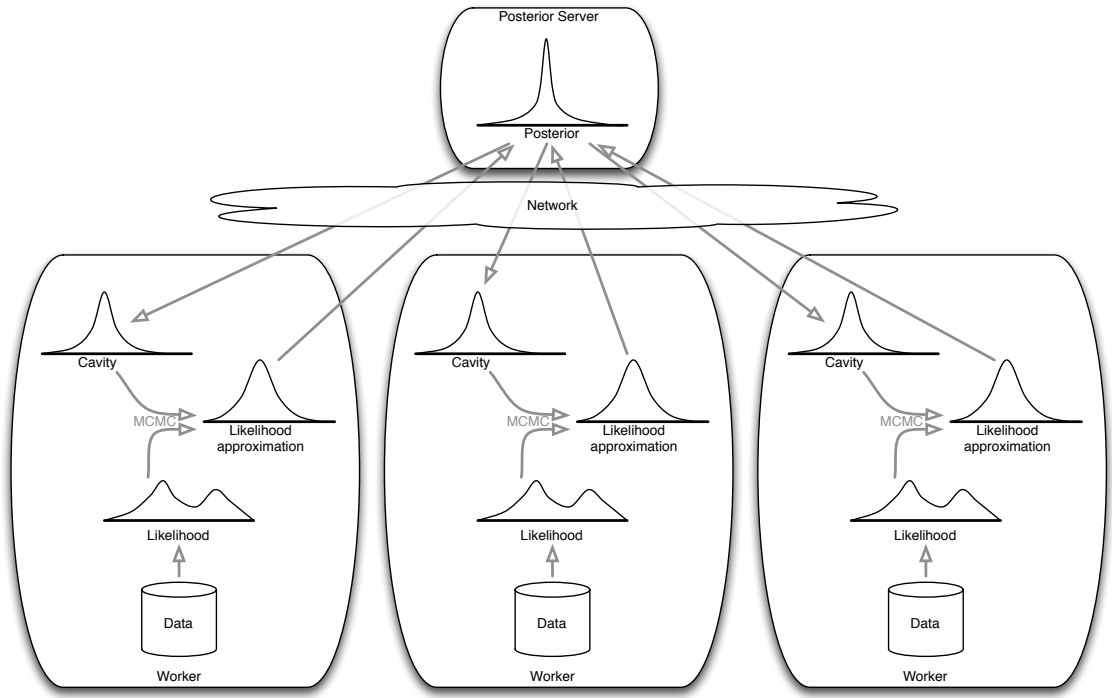

Figure 1: The posterior server.

In the following, Section 2 describes our set up of distributed Bayesian learning and reviews the necessary background on exponential families and convex duality. Section 3 formulates EP and power EP within the framework of variational inference. Our contributions are contained in Section 4 deriving SNEP and Section 5 describing the posterior server architecture. Section B describes additional techniques we used to make the method work on more complex problems like neural networks. We demonstrate the approach Bayesian logistic regression and Bayesian neural networks in Section 6. Section 7 concludes with a summary and discussion of future work.

## 2. Problem Set-up and Background

In this section we set-up the problem of distributed Bayesian learning, using the framework of variational inference in exponential families. For an excellent introduction to exponential families and variational inference we refer the interested reader to (Wainwright and Jordan, 2008). Section 2.1 reviews exponential families and convex duality while introducing nota-tion used throughout the paper. Readers familiar with these concepts can skim ahead to section 2.2.

## 2.1 Exponential Families and Convex Duality

.

Consider an exponential family described by a $d$-dimensional sufficient statistics function $s(x)$. A member $p_\theta$ of this exponential family is parameterized by a natural parameter $\theta \in \mathbb{R}^d$, and has density (with respect to some base measure, say Lebesgue),

$$p_\theta(x) = \exp\left(\theta^\top s(x) - A(\theta)\right),$$
$$A(\theta) = \log \int \exp\left(\theta^\top s(x)\right) dx.$$

The log partition function $A(\theta)$ is convex and finite on the natural domain of the exponential family,

$$\Theta := \{\theta : A(\theta) < \infty\} \subset \mathbb{R}^d,$$

which is a convex subset of $\mathbb{R}^d$.

Associated with any distribution $p$ and the $d$-dimensional sufficient statistics function $s(x)$ is a mean parameter,

$$\mu := \mathbb{E}_p[s(x)],$$

where $\mathbb{E}_p$ denotes the expectation operator with respect to $p$. The set of valid mean parameters $\mathcal{M}$ is a closed convex set, which we refer to as the mean domain,

$$\mathcal{M} = \{\mu : \exists \text{ distribution } p \text{ with } \mu = \mathbb{E}_p[s(x)]\} \subset \mathbb{R}^d$$

Given a natural parameter $\theta \in \Theta$, the exponential family member $p_\theta$ is also associated with a mean parameter $\mu = \mathbb{E}_\theta[s(x)]$ (where $\mathbb{E}_\theta$ denotes the expectation with respect to $p_\theta$), which we can write as a function of the natural parameters, $\mu(\theta)$. If the exponential family is minimal[1], then the mapping $\theta \mapsto \nabla A(\theta)$ is one-to-one and onto the interior of $\mathcal{M}$, and maps $\theta$ to the mean parameter, $\mu(\theta) = \nabla A(\theta)$.

We will assume that our exponential family of interest is minimal.

The convex conjugate of $A(\theta)$ is,

$$A^*(\mu) := \sup_{\theta \in \Theta} \theta^\top \mu - A(\theta).$$

Evaluated at the mean parameter $\mu(\theta)$, the conjugate is the negative entropy of $p_\theta$,

$$A^*(\mu(\theta)) = \mathbb{E}_\theta[\log p_\theta(x)].$$

---

1. The exponential family is minimal if the $d$ 1-dimensional functions making up the sufficient statistics function $s(x)$ are linearly independent, i.e. $\theta^\top s(x) = 0$ for all $x$ implies $\theta = 0$.

Conversely, we have

$$A(\theta) = \sup_{\mu \in \mathcal{M}} \theta^\top \mu - A^*(\mu)$$

and that the natural parameter associated with $\mu$ is $\theta = \theta(\mu) = \nabla A^*(\mu)$. In the following we will make extensive use of the duality between $A$ and $A^*$ and between $\theta$ and $\mu$ described above.

It is useful to write down formulae for the KL divergence between two exponential family distributions, parameterized by natural and mean parameter pairs $\theta, \mu$ and $\theta', \mu'$ respectively:

$$
\begin{aligned}
\mathrm{KL}(p_\theta \| p_{\theta'}) &= \mathbb{E}_\theta[\log p_\theta(x) - \log p_{\theta'}(x)] \\
&= \mathbb{E}_\theta[\theta^\top s(x) - A(\theta) - (\theta')^\top s(x) + A(\theta')] \\
&= \mu^\top(\theta - \theta') - A(\theta) + A(\theta') \\
&= A^*(\mu) + A(\theta') - \mu^\top \theta' \\
&= A^*(\mu) - A^*(\mu') + (\mu' - \mu)^\top \theta'. \quad (1)
\end{aligned}
$$

We will write $\mathrm{KL}(\theta \| \theta'), \mathrm{KL}(\mu \| \theta')$ etc. to refer to the same KL divergence between the same two distributions.

As an example, for a diagonal covariance Gaussian of dimension $d/2$, we have

$$
\begin{aligned}
p(x) &= \prod_{j=1}^{d/2} \frac{1}{\sqrt{2\pi\sigma_j^2}} \exp\left(-\frac{1}{2\sigma_j^2}(x_j - u_j)^2\right) \\
&= \exp\left(\sum_{j=1}^{d/2}(u_j\sigma_j^{-2})(x_j) + (-\sigma_j^{-2})(\tfrac{1}{2}x_j^2) - \tfrac{1}{2}(u_j^2\sigma_j^{-2} + \log(2\pi\sigma_j^2))\right)
\end{aligned}
$$

So the sufficient statistics are $x_j$ and $\frac{1}{2}x_j^2$, mean parameters are $\mu_{j1} = u_j$ and $\mu_{j2} = \frac{1}{2}(u_j^2 + \sigma_j^2)$, natural parameters are $\theta_{j1} = u_j\sigma_j^{-2}$ and $\theta_{j2} = -\sigma_j^{-2}$, and

$$A(\theta) = \sum_{j=1}^{d/2} \frac{1}{2}(u_j^2\sigma_j^{-2} + \log(2\pi\sigma_j^2))$$

$$A^*(\mu) = \sum_{j=1}^{d/2} \frac{-1}{2}(1 + \log(2\pi\sigma_j^2))$$

The conversions between natural and mean parameters are:

$$u_j = \mu_{j1} = -\theta_{j1}\theta_{j2}^{-1} \qquad \theta_{j1} = \mu_{j1}(2\mu_{j2} - \mu_{j1}^2)^{-1} \qquad \mu_{j1} = -\theta_{j1}\theta_{j2}^{-1}$$

$$\sigma_j^2 = 2(\mu_{j2} - \mu_{j1}^2) = -\theta_{j2}^{-1} \qquad \theta_{j2} = -(2\mu_{j2} - \mu_{j1}^2)^{-1} \qquad \mu_{j2} = \frac{1}{2}(\theta_{j1}^2\theta_{j2}^{-2} - \theta_{j2}^{-1})$$

We will use a diagonal covariance Gaussian as our exponential family in most of our experiments, due to the high-dimensionality of the models used.

## 2.2 Distributed Bayesian Learning

We assume that our model is parameterized by a high-dimensional parameter vector $x$. Let the prior distribution $p_0(x)$ be a member of a tractable and minimal exponential family distribution, with natural parameter $\theta_0 \in \Theta \subset \mathbb{R}^d$, sufficient statistics function $s(x)$ and log partition function $A(\theta_0)$. Specifically, we will take $p_0(x)$ to be a diagonal covariance Gaussian. We refer to this exponential family as the *base exponential family*.

We assume that our training data set is spread across a cluster of $n$ compute nodes or workers, with log likelihood $\ell_i(x)$ on compute node $i = 1, \ldots, n$. For example, if we let $\{S_i\}$ be a partition of the data indices, each compute node $i$ could store the corresponding subset of the data $D_i = \{y_c\}_{c \in S_i}$, so that the log likelihood $\ell_i(x)$ is a sum over terms, each corresponding to the log density of one data point stored on node $i$,

$$\ell_i(x) = \sum_{c \in S_i} \log p(y_c \mid x).$$

Covariates and input vectors can be easily incorporated in the above. The target posterior distribution is then,

$$\tilde{p}(x) := p(x \mid \{D_i\}_{i=1}^n) \propto p_0(x) \exp\left(\sum_{i=1}^n \ell_i(x)\right). \qquad (2)$$

Using neural networks as an example, $x$ corresponds to all learnable weights and biases in a network, $\log p(y_c \mid x)$ gives the probability of the class of data item $c$ given the corresponding input vector, and the Gaussian prior corresponds to weight decay.

The learning task is then to compute the posterior distribution. For example, we may want to estimate the posterior mean or variance of the model parameters $x$, or we may want to draw samples distributed according to $\tilde{p}(x)$, using these to predict on test data by averaging the predictive densities over the samples as a Monte Carlo estimate of the marginal predictive density. In the rest of the paper we will aim to obtain these efficiently but approximately.

## 3. Variational Inference in an Extended Exponential Family

In general, the likelihood functions are intractable and approximations are necessary. In this paper we will formulate the learning task as variational inference in an extended exponential family. In particular, we will consider a class of variational methods known as *power expectation propagation* (power EP) (Minka, 2004).

### 3.1 Extended Exponential Family

To start with, we may trivially formulate the target posterior distribution as an *extended exponential family distribution* with sufficient statistics $\tilde{s}(x) := [s(x), \ell_1(x), \ldots, \ell_n(x)]$ and natural parameters $\tilde{\theta} := [\theta_0, \mathbf{1}_n]$ where $\mathbf{1}_n$ is a vector of 1's of length $n$:

$$\tilde{p}(x) = \exp\left(\tilde{\theta}^\top \tilde{s}(x) - \tilde{A}(\tilde{\theta})\right).$$

The extended log partition function $\tilde{A}(\tilde{\theta})$ is (up to a constant) the log marginal probability of the data,

$$
\begin{aligned}
\tilde{A}(\tilde{\theta}) = \log \int \exp\left(\tilde{\theta}^\top \tilde{s}(x)\right) dx &= \log \int \exp\left(\theta_0^\top s(x) + \sum_{i=1}^n \ell_i(x)\right) dx \\
&= \log \mathbb{E}_{\theta_0}\left[\exp\left(\sum_{i=1}^n \ell_i(x)\right)\right] + A(\theta_0) \\
&= \log p(\{D_i\}_{i=1}^n) + A(\theta_0).
\end{aligned}
$$

Denoting the convex conjugate by $\tilde{A}^*(\tilde{\mu})$ and the extended mean domain by $\tilde{\mathcal{M}} \subset \mathbb{R}^{d+n}$, the problem of posterior computation can be expressed as the following concave variational maximization problem:

$$
\max_{\tilde{\mu} \in \tilde{\mathcal{M}}} \tilde{\theta}^\top \tilde{\mu} - \tilde{A}^*(\tilde{\mu}). \tag{3}
$$

For example, if the prior exponential family is a diagonal covariance Gaussian, then the optimal mean parameter is $\tilde{\mu}^* := [\mu^*, \nu_1^*, \ldots \nu_n^*]$, where $\mu^* := \mathbb{E}_{\hat{\theta}}[s(x)] \in \mathbb{R}^d$ corresponds to the posterior means and variances of the model parameters $x$, while $\nu_i^* := \mathbb{E}_{\hat{\theta}}[\ell_i(x)]$ is the posterior expectation of the $i$th log likelihood $\ell_i(x)$. Hence the extended mean parameters capture the important aspects of the posterior distribution.

## 3.2 Power Expectation Propagation

In this section we derive another class of variational approximations corresponding to a generalisation of expectation propagation (EP) (Minka, 2001) called power EP (Minka, 2004). The derivation is a straightforward generalisation of the variational formulation of Wainwright and Jordan (2008) from EP to power EP.

For each worker node $i$ let $\beta_i > 0$ be a given positive real number. Typically, we take $\beta_i = 1$ which corresponds to standard EP, while $\beta_i \to \infty$ corresponds to variational Bayes (Wiegerinck and Heskes, 2003, Minka, 2004). In the formulation of Wainwright and Jordan (2008), EP involves two approximations; both associated with simpler exponential families, which we refer to as *locally extended exponential families* (or sometimes *local exponential families*). For each $i$, let the $i$th locally extended exponential family be associated with the sufficient statistics function $s_i(x) := [s(x), \ell_i(x)]$. Let $\Theta_i$, $\mathcal{M}_i$, $A_i$, $A_i^*$ be the associated (local) natural domain, mean domain, log partition function and negative entropy respectively. A distribution in this locally extended exponential family with natural parameter $[\theta_i, \eta_i] \in \Theta_i$ has the form

$$
p_{\theta_i, \eta_i}(x) = \exp\left(\theta_i^\top s(x) + \eta_i \ell_i(x) - A_i(\theta, \eta_i)\right), \tag{4}
$$

which is a distribution obtained by tilting a tractable distribution with density proportional to $\exp(\theta_i^\top s(x))$ by a single intractable likelihood $\exp(\ell_i(x))$ raised to the power of $\eta_i$. This family can be thought of as treating the $i$th likelihood term exactly, while approximating all other likelihood terms by projecting them onto the tractable base exponential family, the hope being that this family is still tractable while being closer to the true posterior distribution $\tilde{p}(x)$.

For the first approximation, the extended negative entropy $\tilde{A}^*$ is approximated using a tree-like approximation constructed using only the locally extended negative entropies,

$$\tilde{A}^*([\mu, \nu_1, \ldots, \nu_n]) \approx A^*(\mu) + \sum_{i=1}^n \beta_i(A_i^*(\mu, \nu_i) - A^*(\mu)).$$

This approximation is related to the Bethe entropy of loopy belief propagation (Yedidia et al., 2001) and fractional belief propagation (Wiegerinck and Heskes, 2003). Secondly, the extended mean domain $\tilde{\mathcal{M}}$ is approximated by a local mean domain,

$$\mathcal{L} := \{[\mu, \nu_1, \ldots, \nu_n] : [\mu, \nu_i] \in \mathcal{M}_i \text{ for all } i = 1, \ldots, n\}. \tag{5}$$

The local mean domain is an outer bound, $\mathcal{L} \supset \tilde{\mathcal{M}}$. Intuitively, the constraints described by $\tilde{\mathcal{M}}$ are replaced by the weaker constraints described by the local mean domains. We assume that working with the local exponential families in these ways will be more tractable than working with the full extended exponential family.

Let $\mu_0 \in \mathcal{M}$ be a mean parameter in the base exponential family. For the $i$th local exponential family, we denote a mean parameter by $[\mu_i, \nu_i] \in \mathcal{M}_i$, and require the marginalization constraint $\mu_i = \mu_0$. The power EP variational problem, which is not in general concave, is,

$$\begin{aligned} \max_{\mu_0, [\mu_i, \nu_i]_{i=1}^n} \quad & \theta_0^\top \mu_0 + \sum_{i=1}^n 1 \cdot \nu_i - A^*(\mu_0) - \sum_{i=1}^n \beta_i(A_i^*(\mu_i, \nu_i) - A^*(\mu_i)) \\ \text{subject to} \quad & \mu_0 \in \mathcal{M} \\ & [\mu_i, \nu_i] \in \mathcal{M}_i \quad \text{for } i = 1, \ldots, n \\ & \mu_0 = \mu_i \quad \text{for } i = 1, \ldots, n \end{aligned} \tag{6}$$

Note in particular that the entropy and mean domain in the original variational problem (3) have been replaced by their respective approximations.

### 3.3 Deriving EP and Power EP Updates

In this section, for completeness' sake, we derive the updates for EP and power EP as fixed-point equations that solve the variational problem. Readers familiar with this derivation can skip ahead to section 3.4.

First we introduce Lagrange multipliers $\lambda_i$ for the equality constraints $\mu_0 = \mu_i$, so that the above is equivalent to,

$$\begin{aligned} \max_{\mu_0, [\mu_i, \nu_i]_{i=1}^n} \min_{[\lambda_i]_{i=1}^n} \quad & \underbrace{\theta_0^\top \mu_0 - A^*(\mu_0) + \sum_{i=1}^n \left(\nu_i - \lambda_i^\top(\mu_i - \mu_0) - \beta_i(A_i^*(\mu_i, \nu_i) - A^*(\mu_i))\right)}_{=:L(\mu_0, [\mu_i, \nu_i, \lambda_i]_{i=1}^n)} \\ \text{subject to} \quad & \mu_0 \in \mathcal{M} \\ & [\mu_i, \nu_i] \in \mathcal{M}_i \quad \text{for } i = 1, \ldots, n \end{aligned} \tag{7}$$

where the domain of $\lambda_i$ is $\mathbb{R}^d$. Let the Lagrangian above be denoted by $L(\mu_0, [\mu_i, \nu_i, \lambda_i]_{i=1}^n)$. The Karush-Kuhn-Tucker (KKT) conditions of the above variational problem have to be satisfied at an optimum, and simply involve setting the derivatives with respect to $\mu_0, \mu_i, \nu_i, \lambda_i$ to zero:

$$\frac{dL}{d\lambda_i} = 0 \; : \qquad\qquad\qquad \mu_i = \mu_0 \qquad\qquad (8a)$$

$$\frac{dL}{d\mu_0} = 0 \; : \qquad\qquad \theta_0 - \nabla A^*(\mu_0) + \sum_{j=1}^n \lambda_j = 0$$

$$\theta_0 + \sum_{j=1}^n \lambda_j = \nabla A^*(\mu_0) \qquad\qquad (8b)$$

$$\frac{dL}{d\mu_i} = 0 \; : \qquad -\lambda_i - \beta_i \nabla_{\mu_i} A_i^*(\mu_i, \nu_i) + \beta_i \nabla A^*(\mu_i) = 0$$

$$\theta_0 + \sum_{j=1}^n \lambda_j - \beta_i^{-1} \lambda_i = \nabla_{\mu_i} A_i^*(\mu_i, \nu_i) \qquad\qquad (8c)$$

$$\frac{dL}{d\nu_i} = 0 \; : \qquad\qquad\qquad \beta_i^{-1} = \nabla_{\nu_i} A_i^*(\mu_i, \nu_i) \qquad\qquad (8d)$$

Equation (8b) shows that an optimal $\mu_0$ has to be the mean parameter corresponding to a (base) exponential family distribution with natural parameter $\theta_0 + \sum_{j=1}^n \lambda_j$. Specifically, the posterior distribution can be approximated as,

$$\tilde{p}(x) \propto p_0(x) \exp\left(\sum_{j=1}^n \ell_j(x)\right) \approx \exp\left(\left(\theta_0 + \sum_{j=1}^n \lambda_j\right)^\top s(x)\right). \qquad (9)$$

In other words, we can interpret $\lambda_j$ as the natural parameter of an exponential family approximation to the likelihood factor $\exp(\ell_j(x))$.

Further, from (8c) and (8d) above, we see that the optimal $[\mu_i, \nu_i]$ is the mean parameter associated with the local posterior distribution,

$$p_i(x) \propto \exp\left(\left(\theta_0 + \sum_{j=1}^n \lambda_j - \beta_i^{-1} \lambda_i\right)^\top s(x) + \beta_i^{-1} \ell_i(x)\right). \qquad (10)$$

For standard EP, where $\beta_i = 1$, we get,

$$p_i(x) \propto \exp\left(\left(\theta_0 + \sum_{j\neq i} \lambda_j\right)^\top s(x) + \ell_i(x)\right).$$

The above local posterior distribution is known as the tilted distribution in EP, with the term $(\theta_0 + \sum_{j=1}^n \lambda_j - \beta_i^{-1} \lambda_i)^\top s(x)$ corresponding to the cavity distribution with (the $\beta_i^{-1}$th power of) the exponential family approximation to the $i$th likelihood removed, and replaced by (the $\beta_i^{-1}$th power of) the likelihood factor itself. Each step of EP involves first computing $[\mu_i, \nu_i]$ as the mean parameter of the tilted distribution, then updating $\lambda_i$, using (8a) and (8b):

$$\lambda_i^{\text{new}} = \nabla A^*(\mu_i) - \theta_0 - \sum_{j \neq i} \lambda_j, \qquad (11)$$

where we recall that $\nabla A^*(\mu_i)$ computes the natural parameter corresponding to the mean parameter $\mu_i$ in the base exponential family. The EP update ensures that the expectation of the sufficient statistics function under the tilted distribution $p_i(x)$ and under its exponential family approximation (with natural parameters $\theta_0 + \sum_{j=1}^n \lambda_j - \beta_i^{-1}\lambda_i + \beta_i^{-1}\lambda_i^{\text{new}}$) match. At convergence, this ensures that the expectations under all the tilted distributions and under the approximated posterior distribution (9) match.

Note that the (power) EP updates are derived as fixed point equations and have no guarantees of convergence. In practice, damped updates can be used to avoid oscillations without affecting the fixed points:

$$\lambda_i^{\text{new}} = \alpha\lambda_i + (1 - \alpha)\left(\nabla A^*(\mu_i) - \theta_0 - \sum_{j \neq i} \lambda_j\right). \qquad (12)$$

where $\alpha \in [0, 1)$ is a damping factor. Another potential issue is that $\lambda_i^{\text{new}}$ as computed may not lie in the natural domain $\Theta$, in which case higher damping factors ensuring that it does can be used, or the update can be discarded.

### 3.4 Computing Mean Parameters

Assuming that the base exponential family is tractable, the above steps of EP involve additions, subtractions, and conversions between mean and natural parameters so are easy to compute. The only difficulty is in the computation of the mean parameters (expectations of the sufficient statistics function) of the tilted distributions, which involve the log likelihoods $\ell_i(x)$. In typical applications of EP to graphical models, these are either obtained analytically or using numerical quadrature.

In more complex and general scenarios, both analytic or quadrature-based methods are ruled out. A successful and common class of methods for calculating expectations under otherwise intractable distributions is Monte Carlo. A number of papers have proposed such an approach, including Barthelmé and Chopin (2011), Heess et al. (2013), Gelman et al. (2014) using importance sampling and Xu et al. (2014), Gelman et al. (2014) using Markov chain Monte Carlo (MCMC). In addition, Heess et al. (2013), Eslami et al. (2014), Jitkrittum et al. (2015) use learning techniques to speed-up the process by directly predicting the natural parameters given properties of the tilted distribution.

We have explored the sampling via moment sharing (SMS) algorithm (Xu et al., 2014) in the context of distributed Bayesian learning of deep neural networks. Unfortunately, the SMS algorithm did not work even for relatively small neural networks and data sets, partly because of the high dimensionality and partly because the Monte Carlo estimation is inherently stochastic, both of which we found affected the convergence of the EP fixed point equations.

## 4. Stochastic Natural-gradient Expectation Propagation

In this section we will derive a novel convergent stochastic approximation based alternative to EP and power EP, which optimises the same variational objective but is significantly more tolerant of Monte Carlo noise. Our algorithm is derived using a modified variational objective with additional auxiliary variables but with the same optima as the original problem, and solving the dual problem using a stochastic approximation algorithm (Robbins and Monro, 1951). In the following we will use "EP" to refer to both EP and power EP for simplicity.

### 4.1 Auxiliary Variational Problem

For each $i$, we introduce an auxiliary natural parameter vector $\theta_i' \in \Theta$, and introduce a term $-\sum_{i=1}^n \mathrm{KL}(p_{\mu_i} \| p_{\theta_i'})$ into the variational objective (7), where $p_{\mu_i}$ and $p_{\theta_i'}$ are the base exponential family distribution given by mean parameter $\mu_i$ and natural parameter $\theta_i'$, respectively. This results in a lower bound on the original objective and is reminiscent of the EM algorithm (Dempster et al., 1977, Neal and Hinton, 1999) and of typical variational Bayes approximations (Beal, 2003, Wainwright and Jordan, 2008). Plugging the relevant formula for the KL divergence (1) into (7), we have the resulting variational problem

$$
\begin{aligned}
\max_{\mu_0, [\mu_i, \nu_i, \theta_i']_{i=1}^n} \quad & \theta_0^\top \mu_0 - A^*(\mu_0) + \sum_{i=1}^n \left( \nu_i - \beta_i \left( A_i^*(\mu_i, \nu_i) - \mu_i^\top \theta_i' + A(\theta_i') \right) \right) \\
\text{subject to} \quad & \mu_0 \in \mathcal{M} \\
& [\mu_i, \nu_i] \in \mathcal{M}_i \quad \text{for } i = 1, \ldots, n \\
& \theta_i' \in \Theta \quad \text{for } i = 1, \ldots, n \\
& \mu_0 = \mu_i \quad \text{for } i = 1, \ldots, n
\end{aligned}
\tag{13}
$$

Maximizing over $\theta_i'$ while keeping the other variables fixed will simply set $\theta_i' = \nabla A^*(\mu_i)$, so that the KL terms vanish and resulting in the original problem (6). On the other hand, the constrained maximization over the original parameters $\mu_0, [\mu_i, \nu_i]_{i=1}^n$ requires introducing Lagrange multipliers,

$$
\begin{aligned}
\max_{[\theta_i']_{i=1}^n} \max_{\mu_0, [\mu_i, \nu_i]_{i=1}^n} \min_{[\lambda_i]_{i=1}^n} \quad & \theta_0^\top \mu_0 - A^*(\mu_0) + \sum_{i=1}^n \left( \nu_i - \lambda_i^\top (\mu_i - \mu_0) - \beta_i \left( A_i^*(\mu_i, \nu_i) - \mu_i^\top \theta_i' + A(\theta_i') \right) \right) \\
\text{subject to} \quad & \mu_0 \in \mathcal{M} \\
& [\mu_i, \nu_i] \in \mathcal{M}_i \quad \text{for } i = 1, \ldots, n \\
& \theta_i' \in \Theta \quad \text{for } i = 1, \ldots, n
\end{aligned}
\tag{14}
$$

We can solve this inner constrained optimization by transforming to the convex dual. Noticing that the Lagrangian is concave in $\mu_0, [\mu_i, \nu_i]_{i=1}^n$ and that Slater's condition holds, the duality gap is zero and we have,

$$\max_{[\theta_i']_{i=1}^n} \min_{[\lambda_i]_{i=1}^n} \max_{\mu_0, [\mu_i, \nu_i]_{i=1}^n} \theta_0^\top \mu_0 - A^*(\mu_0) + \sum_{i=1}^n \left( \nu_i - \lambda_i^\top (\mu_i - \mu_0) - \beta_i \left( A_i^*(\mu_i, \nu_i) - \mu_i^\top \theta_i' + A(\theta_i') \right) \right)$$

$$\text{subject to} \quad \mu_0 \in \mathcal{M} \tag{15}$$
$$[\mu_i, \nu_i] \in \mathcal{M}_i \quad \text{for } i = 1, \dots, n$$
$$\theta_i' \in \Theta \quad \text{for } i = 1, \dots, n$$

Now maximizing over $\mu_0, [\mu_i, \nu_i]_{i=1}^n$, we have the equivalent dual problem, whose objective function we will denote with $L([\theta_j', \lambda_j]_{j=1}^n)$,

$$\max_{[\theta_i']_{i=1}^n} \min_{[\lambda_i]_{i=1}^n} \underbrace{A \left( \theta_0 + \sum_{i=1}^n \lambda_i \right) + \sum_{i=1}^n \beta_i \left( A_i \left( \theta_i' - \beta_i^{-1} \lambda_i, \beta_i^{-1} \right) - A(\theta_i') \right)}_{=: L([\theta_j', \lambda_j]_{j=1}^n)}$$

$$\text{subject to} \quad \theta_i' \in \Theta \quad \text{for } i = 1, \dots, n \tag{16}$$

In summary, our algorithm is coordinate maximization of (13), where we alternate between maximizing $[\theta_i']_{i=1}^n$ given $\mu_0, [\mu_i, \nu_i]_{i=1}^n$, and maximizing $\mu_0, [\mu_i, \nu_i]_{i=1}^n$ given $[\theta_i']_{i=1}^n$. To solve the second maximization, we transform to the dual and minimize (16) with respect to the Lagrange multipliers $[\lambda_i]_{i=1}^n$. This requires an inner iteration loop involving stochastic natural gradient descent, which is described next. The structure of our derived algorithm is the same as the convergent double loop EP algorithm of (Heskes and Zoeter, 2002). Our derivation is interesting and useful in that a single global variational problem is described, rather than a sequence of variational problems each obtained by minorizing around the current parameters and then maximizing, as in the minorization-maximization (MM) algorithm) (Hunter and Lange, 2004) and the CCCP algorithm (Yuille, 2002). Our algorithm is also different in the choice of stochastic natural gradient descent inner loop.

## 4.2 Stochastic Natural Gradient Descent

The gradient of the dual objective $L([\theta_j', \lambda_j]_{j=1}^n)$ is,

$$\frac{dL}{d\lambda_i} = \nabla A \left( \theta_0 + \sum_{j=1}^n \lambda_j \right) - \nabla_{\theta_i} A_i \left( \theta_i' - \beta_i^{-1} \lambda_i, \beta_i^{-1} \right) \tag{17}$$

where we used the notation $\nabla_{\theta_i} A_i$ for the partial derivative of $A_i(\cdot, \cdot)$ with respect to its first ($d$-dimensional) argument. While this gradient is the direction of steepest descent in a Euclidean geometry it does not take into account the geometry of the base exponential family. In other words, the gradient is not covariant[2] (MacKay, 1999) and the corresponding gradient descent algorithm is not expected to perform well.

---

2. A covariant gradient descent algorithm roughly means that the size of the gradient in a direction positively covaries with the length scale in that direction hence with the desired step size. An example of a non-covariant gradient descent algorithm is steepest descent with standard Euclidean geometry, say for a 2D quadratic loss function that is long in the $x$-axis and short in the $y$-axis. The gradient will be close to pointing vertically up or down, but the best descent direction points horizontally instead.

A better approach is to use natural gradient descent (Amari et al., 1996, Amari and Nagaoka, 2001) or, equivalently, mirror descent (Beck and Teboulle, 2003, Raskutti and Mukherjee, 2015) instead. These methods treat the parameter space as a Riemannian manifold where the metric tensor is given by the Fisher information. The direction of steepest descent in this geometry is given by the gradient preconditioned by the inverse metric. Natural gradient descent has appealing theoretical properties as it incorporates second order information to be more covariant. As we will see below we can take advantage of natural gradient descent at no additional computational cost by reparametrising. As noted in the previous section, the Lagrange multiplier $\lambda_i$ can be interpreted as the natural parameters of the base exponential family approximation to the likelihood $\exp(\ell_i(x))$. Reparameterising using the corresponding mean parameter $\gamma_i$ instead, with $\lambda_i = \nabla A^*(\gamma_i)$, the gradient is,

$$
\begin{aligned}
\frac{dL}{d\gamma_i} &= \frac{d\lambda_i}{d\gamma_i} \left( \nabla A \left( \theta_0 + \sum_{j=1}^n \lambda_j \right) - \nabla_{\theta_i} A_i \left( \theta_i' - \beta_i^{-1}\lambda_i, \beta_i^{-1} \right) \right) \\
&= \nabla^2 A^*(\gamma_i) \left( \nabla A \left( \theta_0 + \sum_{j=1}^n \lambda_j \right) - \nabla_{\theta_i} A_i \left( \theta_i' - \beta_i^{-1}\lambda_i, \beta_i^{-1} \right) \right)
\end{aligned}
\tag{18}
$$

The appropriate metric in the mean parameter space is simply $\nabla^2 A^*(\gamma_i)$, so that its inverse cancels the $\nabla^2 A^*(\gamma_i)$ term (Raskutti and Mukherjee, 2015), and the natural gradient update is simply,

$$
\gamma_i^{(t+1)} = \gamma_i^{(t)} + \epsilon_t \left( \nabla_{\theta_i} A_i \left( \theta_i' - \beta_i^{-1}\lambda_i^{(t)}, \beta_i^{-1} \right) - \nabla A \left( \theta_0 + \sum_{j=1}^n \lambda_j^{(t)} \right) \right)
\tag{19}
$$

where the corresponding natural parameter is given as a function of the mean parameter, $\lambda_i^{(t)} := \nabla A^*(\gamma_i^{(t)})$, and $\epsilon_t$ is the step size at iteration $t$. These updates can be performed in series or parallel fashion. In our distributed Bayesian learning setting they are performed in an asynchronous distributed fashion (Section 5).

Recall that derivatives of the log partition function compute the mean parameters from the natural parameters. Hence the first term of the natural gradient step is the mean parameter of the current local tilted distribution,

$$
p_i^{(t)}(x) = \exp \left( (\theta_i' - \beta_i^{-1}\lambda_i^{(t)})^\top s(x) + \beta_i^{-1}\ell_i(x) - A_i(\theta_i' - \beta_i^{-1}\lambda_i^{(t)}, \beta_i^{-1}) \right),
\tag{20}
$$

while the second term is the mean parameter of the current exponential family approximation to the global posterior. Their difference gives the update for the mean parameter of the likelihood approximation. The gradient is zero when both terms are equal, precisely the condition from which the EP fixed point equation is derived.

In general, the first term cannot be obtained in closed form, and we instead use a Markov chain Monte Carlo (MCMC) estimate, leading to a stochastic natural-gradient descent algorithm. Specifically, let $\mathcal{K}_i(\cdot \mid x; \theta_i' - \beta_i^{-1}\lambda_i^{(t)}, \beta_i)$ be a Markov chain kernel with previous state $x$ whose invariant distribution is the local tilted distribution (20). Let $x_i^{(t)}$ be the state of the Markov chain at iteration $t$. The next state $x_i^{(t+1)}$ is obtained by simulating from the Markov chain, using the current values of the parameters,

$$x_i^{(t+1)} \sim \mathcal{K}_i \left( \cdot \mid x_i^{(t)}; \theta_i' - \beta_i^{-1} \lambda_i^{(t)}, \beta_i \right). \tag{21}$$

In summary, the stochastic natural gradient update is,

$$\gamma_i^{(t+1)} = \gamma_i^{(t)} + \epsilon_t \left( s \left( x_i^{(t+1)} \right) - \nabla A \left( \theta_0 + \sum_{j=1}^n \lambda_j^{(t)} \right) \right) \tag{22}$$

We refer to the resulting algorithm as "Stochastic Natural-gradient EP", or SNEP for short.

Technically, the stochastic natural-gradient descent requires unbiased estimates of gradients, and the mean parameter estimates obtained using MCMC updates are only unbiased if the Markov chain equilibrates in between gradient updates. In practice, we find that using MCMC samples works well regardless, an observation consistent with past experiences with stochastic approximation algorithms in machine learning (Neal and Hinton, 1999, Tieleman, 2008).

The stochastic natural gradient descent algorithm above serves as the inner loop of our double loop algorithm, and solves for the constrained maximization of the mean parameters $\mu_0, [\mu_i, \nu_i]_{i=1}^n$ given auxiliary parameters $[\theta_i']_{i=1}^n$. Returning now to the outer loop update, maximizing $\theta_i'$ involves simply setting it to be $\nabla A^*(\mu_i)$, the natural parameter corresponding to the mean parameter $\mu_i$ of the local tilted distribution (see the discussion after Equation 13). Assuming that the inner loop has converged, this and the mean parameters of all other tilted distributions would be equal to $\mu_0$, so that the $t'$th outer loop update is,

$$(\theta_i')^{(t'+1)} = \nabla A^*(\mu_i) = \nabla A^*(\mu_0) = \theta_0 + \sum_{j=1}^n \lambda_j^{(\infty)} \tag{23}$$

where $\lambda_j^{(\infty)}$ is the converged value of the dual parameters in the inner loop. In practice, we simply perform the outer loop update infrequently (and before the inner loop has fully converged). In our experiments we do not see instabilities resulting from this, an observation also consistent with past experiences with double loop algorithms (Teh and Welling, 2002, Yuille, 2002, Heskes and Zoeter, 2002).

In the extreme case where the outer loop update is performed after every inner loop update, we can roll both updates into the following update,

$$\gamma_i^{(t+1)} = \gamma_i^{(t)} + \epsilon_t \left( \nabla_{\theta_i} A_i \left( \theta_0 + \sum_{j=1}^n \lambda_j^{(t)} - \beta_i^{-1} \lambda_i^{(t)}, \beta_i^{-1} \right) - \nabla A \left( \theta_0 + \sum_{j=1}^n \lambda_j^{(t)} \right) \right) \tag{24}$$

It is interesting to contrast the above with a damped EP update, which in our notation is given by,

$$\lambda_i^{(t+1)} = \lambda_i^{(t)} + \epsilon_t \left( \nabla A^* \left( \nabla_{\theta_i} A_i \left( \theta_0 + \sum_{j=1}^n \lambda_j^{(t)} - \beta_i^{-1} \lambda_i^{(t)}, \beta_i^{-1} \right) \right) - \left( \theta_0 + \sum_{j=1}^n \lambda_j^{(t)} \right) \right) \tag{25}$$

Note that $\nabla A^*(\cdot)$ converts from mean parameters to natural parameters, so that the first term in parentheses can be read as first computing the moments (mean parameters) of the tilted distribution then converting into the corresponding natural parameter. Hence each term in the damped EP update is obtained by converting the corresponding term in (24) from mean to natural parameters, i.e. applying $\nabla A^*(\cdot)$. In other words the update (24) can be thought of as a mean parameter space version of a damped EP update. When mean parameters are estimated with Monte Carlo noise, updating using mean parameters rather than natural parameters leads to more stable behaviour; see Section 6.1 for empirical results supporting this claim.

### 4.3 Discussion and Related Works

While we developed SNEP in the context of distributed Bayesian learning, it is clear that it is generally applicable, since the distribution (2) targeted is simply a product of factors, each of which is approximated by a factor in the base exponential family, precisely the setting of EP. SNEP can therefore be used in place of EP in situations where Monte Carlo moment estimates are used, including graphical models (Heess et al., 2013, Eslami et al., 2014, Jitkrittum et al., 2015, Lienart et al., 2015), hierarchical Bayesian models (Gelman et al., 2014), and approximate Bayesian computation (Barthelmé and Chopin, 2011).

Since Minka (2001), there have been a substantial number of extensions and alternatives to EP proposed. Stochastic EP (Li et al., 2015) and averaged EP (Dehaene and Barthelmé, 2015) assume that all factors can be well approximated by the *same* exponential family factor via parameter tying. This saves memory storage and was shown to work well. This is an orthogonal idea to our work, and it is possible to apply it to SNEP as well in the future, although in the distributed learning setting considered in this paper each worker must keep a copy of the parameters anyway, which means it would not reduce memory requirements. Convergent EP (Heskes and Zoeter, 2002) is also a double loop coordinate descent algorithm with convergence guarantees but SNEP differs in that the inner loop is a stochastic natural-gradient descent which is more robust to the use of Monte Carlo estimated moments. We note here that stochastic natural-gradient descent has also been used in Stochastic Variational Inference (Hoffman et al., 2013).

SNEP can be considered a black-box variational inference algorithm, as the only requirement on the model is the existence of MCMC samplers targeting the tilted distributions. Black-box methods have recently been developed for variational Bayes (Ranganath et al., 2014, Black-box Variational Inference or BBVI) and for power EP (Hernandez-Lobato et al., 2016, Black-box Alpha or BB-$\alpha$). BB-$\alpha$ and SNEP are both methods for Bayesian learning based on power EP and both operate in the mean parameter space. BB-$\alpha$ uses the parameter tying idea of stochastic and averaged EP to construct a further approximated objective function which is then optimised using a convergent single-loop algorithm. As a result the fixed points of the algorithm are different from power EP. Single loop algorithms are generally preferable as it is well known that double loop algorithms can be slow if the inner loop is run until convergence. In our experiments, we do not run the inner loop of SNEP until convergence, and have found empirically that even a single iteration of the inner loop per outer loop update does not affect convergence or stability. In fact, on a two-layer feedforward neural network, SNEP converges in less than one tenth of the number of epochs and a fraction of the running times of those quoted by Hernandez-Lobato et al. (2016). In both BBVI and BB-$\alpha$, naive Monte Carlo estimators are used, with samples drawn from the approximating distribution. The resulting estimators can have high variance, requiring techniques for variance reduction. In contrast, SNEP uses MCMC samplers targeting the tilted distribution. It is generally accepted that, in high-dimensional settings, MCMC samplers often have lower variance than naive Monte Carlo and as a result work better, with the tradeoff being that MCMC samplers need to equilibrate and produce correlated samples.

Our application of SNEP to distributed Bayesian learning applies one exponential family approximation per subset of data on each worker node. This contrasts with typical applications of EP and variational inference in general, which applies one approximation per data item. This is made possible due to the black-box flexibility of our approach, since the likelihood associated with a data subset is more complex than for a single data item. In cases where the subset is itself quite large, and the Bernstein-von Mises theorem holds, the likelihood will be close to a Gaussian, so that if we use a (full-covariance) Gaussian as the base exponential family, the approximation will introduce negligible biases. For a recent study of EP in the large data limit, see (Dehaene and Barthelmé, 2015). We can think of our approach as a hybrid which interpolates between a pure variational approach (when $n = N$) and a pure MCMC approach (when $n = 1$), with smaller $n$ corresponding to less bias introduced by approximations but higher variance/computational costs.

In our experiments, we apply SNEP to deep neural networks. Deep learning has been extremely successful in a large number of different domains. However, one drawback of neural networks trained by stochastic optimisation techniques is that they do not provide uncertainty estimates. Both black-box variational methods and stochastic gradient MCMC methods can be applied to neural networks yielding uncertainty estimates. Probabilistic backpropagation (Hernandez-Lobato and Adams, 2015, PBP) relies on assumed density filtering (ADF), a sequential form of EP. Other approaches are based on mean field variational inference, which optimises a lower bound of the marginal likelihood. Graves (2011) proposed an algorithm using a biased Monte Carlo estimate of this lower bound to optimise it. More recently, this approach has been extended by Blundell et al. (2015, Bayes by Backprop) who obtain unbiased Monte Carlo estimates of the same objective using the reparametrization trick (Kingma and Welling, 2014).

## 5. The Posterior Server

Our development of SNEP is motivated by the problem of distributed Bayesian learning outlined in Section 2.2, where each log likelihood term $\ell_i(x)$ corresponds to the log probability/density of the data subset on worker node $i$. Using SNEP, each worker node iteratively learns an exponential family approximation of $\ell_i(x)$, with a master node coordinating the learning across workers. We call the master node the *posterior server*, as it maintains and serves the exponential family approximation of the posterior distribution, obtained by combining the prior with the likelihood approximations at the workers.

In more detail, the posterior server maintains the natural parameter $\theta_{\text{posterior}} := \theta_0 + \sum_{j=1}^{n} \lambda_j$ of the global posterior approximation, while each worker node $i$ maintains the mean and natural parameters $\gamma_i, \lambda_i$ of the likelihood approximation and the state of the MCMC sampler $x_i$. It also maintains the auxiliary parameter $\theta_i'$ used in the outer loop. Learning at the worker proceeds by alternating between the MCMC (21) and inner loop updates (22), with periodic outer loop updates (23). These updates require access to the data subset at the node, as well as the cavity distribution, with natural parameters $\theta_{-i} := \theta_0 + \sum_{j \neq i} \lambda_j$, which is obtained by getting $\theta_{\text{posterior}}$ from the posterior server and subtracting the local $\lambda_i$.

Communication between the worker node and the posterior server involves the worker first sending the posterior server the difference $\Delta_i := \lambda_i^{\text{new}} - \lambda_i^{\text{old}}$ between the current natural parameters $\lambda_i^{\text{new}}$ and the one during the previous communication with the server, $\lambda_i^{\text{old}}$. The posterior server updates its global posterior approximation via $\theta_{\text{posterior}}^{\text{new}} = \theta_{\text{posterior}}^{\text{old}} + \Delta_i$, and sends the new value $\theta_{\text{posterior}}^{\text{new}}$ back to the worker. The worker in turn uses this to update the cavity, $\theta_{-i}^{\text{new}} = \theta_{\text{posterior}}^{\text{new}} - \lambda_i^{\text{new}}$.

The pseudocode for the overall algorithm is given in Algorithm 1. Note that all communications are performed asynchronously and in a non-blocking fashion. In particular, Steps 11-17 are performed in a separate coroutine from the main loop (Steps 7-18), and Step 15 can happen several iterations of the main loop after Step 14. This is so that compute nodes can spend most of their time learning (Steps 8-10) and do not have to wait for network communications to complete. However, this means that the previous copy of the natural parameter $\lambda_i$ used when the last message was sent to the master (Step 13) needs to be stored, in addition to the most recent one. We also note that faster compute nodes need not wait for slower ones since they each learn their own separate likelihood approximation parameters. It would be interesting for future research to explore adaptive methods to allow faster compute nodes to increase the data subsets that they learn from, and slower ones to decrease, to balance the learning progress across compute nodes more evenly.

## 5.1 Discussion and Related Works

Our naming of the posterior server contrasts with that of the parameter server (Ahmed et al., 2012) which is typically used for maximum likelihood (or minimum empirical risk) estimation of model parameters. Note however that our algorithmic contribution is effectively orthogonal to (Ahmed et al., 2012), who proposed a generic and robust computational architecture for distributed machine learning. We believe it is possible to implement our algorithm using the parameter server software framework.

One of the difficulties of the parameter server architecture is that learning happens at the level of parameters, with a single set of parameters being maintained across the cluster. Since the data subsets on workers are disjoint, the learning on each worker tends to make the local copy of the parameters diverge from those on the parameter server and on other workers. As a result, frequent synchronisation with the parameter server is necessary to prevent stale parameters and gradients. As an example, in the DistBelief method (Dean et al., 2012), experiments were conducted where the communication with the master was performed after every iteration, which can significantly slow down the learning process. On the other hand, one of the interesting aspects of the posterior server is that it lifts learning from the level of parameters to the level of distributions over parameters. As a result each worker can maintain a distinct parameter set in the MCMC sampler and a distinct likelihood approximation (since the likelihoods on different workers are indeed different as they have different data subsets) without requiring frequent communication with the posterior server.

---

**Algorithm 1** Posterior Server: Distributed Bayesian Learning via SNEP

---

1: **for** each compute node $i = 1, \ldots, n$ **asynchronously do**
2:    let $\gamma_i^{(1)}$ be the initial mean parameter of local likelihood approximation.
3:    let $\lambda_i^{\text{old}} := \lambda_i^{(1)} := \nabla A^*(\gamma_i^{(1)})$ be the initial natural parameter of local likelihood approximation.
4:    let $\theta_{-i} := \theta_0 + \sum_{j \neq i} \lambda_j^{(1)}$ be the initial natural parameter of cavity distribution
5:    let $\theta_i' := \theta_{-i} + \lambda_i^{(1)}$ be the initial auxiliary parameter.
6:    let $x_i^{(1)} \sim p_{\theta_{-i} + \lambda_i^{(1)}}$ be the initial state of MCMC sampler.
7:    **for** $t = 1, 2, \ldots$ until convergence **do**
8:       update local state via MCMC:

$$x_i^{(t+1)} \sim \mathcal{K}_i \left( \cdot \mid x_i^{(t)}; \theta_i' - \beta_i^{-1} \lambda_i^{(t)}, \beta_i^{-1} \right)$$

9:       update local likelihood approximation:

$$\gamma_i^{(t+1)} := \gamma_i^{(t)} + \epsilon_t \left( s(x_i^{(t+1)}) - \nabla A \left( \theta_{-i} + \lambda_i^{(t)} \right) \right)$$
$$\lambda_i^{(t+1)} := \nabla A^*(\gamma_i^{(t+1)})$$

10:      **every** $N_{\text{outer}}$ iterations **do**: update auxiliary parameter:

$$\theta_i' := \theta_{-i} + \lambda_i^{(t)}$$

11:      **every** $N_{\text{sync}}$ iterations **asynchronously do:** communicate with posterior server:
12:         let $\Delta_i := \lambda_i^{(t)} - \lambda_i^{\text{old}}$.
13:         update $\lambda_i^{\text{old}} := \lambda_i^{(t)}$.
14:         **send** $\Delta_i$ to posterior server.
15:         **receive** $\theta_{\text{posterior}}$ from posterior server.
16:         update $\theta_{-i} := \theta_{\text{posterior}} - \lambda_i^{\text{old}}$.
17:    **end for**
18: **end for**
19: **for** the posterior server **do**
20:    let $\theta_{\text{posterior}} := \theta_0 + \sum_{j=1}^{n} \lambda_i^{(1)}$ be the initial natural parameter of the posterior approximation.
21:    maintain a queue of messages from workers.
22:    **for** each message $\Delta_i$ received from some worker $i$ **do**
23:       update $\theta_{\text{posterior}} := \theta_{\text{posterior}} + \Delta_i$.
24:       **send** $\theta_{\text{posterior}}$ to worker $i$.
25:    **end for**
26: **end for**

---

The only role of communication here is for the cavity distributions, which can be thought of as a way for the system to focus the learning happening on the workers on the relevant regions of the parameter space (Xu et al., 2014). Empirically, the precise parameterisation of the cavity distribution is not very important. For an extreme example, suppose both the prior and likelihood terms are close to being Gaussians and the tractable family is also Gaussian. Then the likelihood approximation will not in fact depend on the cavity distribution at all, and will converge to the true likelihood on each worker independently. See also Zhang et al. (2015a) for elastic-averaging SGD, a similar idea of allowing each worker a separate parameter vector, using an ADMM-like methodology.

## 6. Experiments

In this section we present our experiments on SNEP and the posterior server. Our implementation, which is available at `http://bigbayes.github.com/PosteriorServer`, is in the Julia[3] technical computing language. In our setup, the master and workers are run on separate system processes (that is, in separate memory spaces) on a many-core computing cluster, making use of the high-level parallel programming abstractions provided by Julia. Other architectures are possible; for example, multiple threads sharing an address space, multiple GPUs, or nodes communicating over a network using MPI. For the MCMC sampler on workers, we chose an adaptive version of stochastic gradient Langevin dynamics (SGLD) (Welling and Teh, 2011) related to Adam (Kingma and Ba, 2015), which is more computationally scalable to larger neural network models and data sets than standard MCMC (see Appendix B for details). Our neural network models are implemented in the Mocha[4] deep learning Julia package.

### 6.1 Comparison to SMS on Bayesian Logistic Regression

In this section we compare SNEP against *Sampling via Moment Sharing* (SMS) (Xu et al., 2014), a related algorithm for distributed Bayesian learning whereby each worker has a separate MCMC sampler and coordination across workers is achieved using plain EP. SMS was originally proposed to scale up MCMC methods and as such it assumes that MCMC chains can be run for many iterations to convergence in between communications with the master. SMS uses the MCMC iterates to estimate the moments required for EP updates and so, as discussed in the introduction, requires a large number of iterations to produce the low Monte Carlo noise for EP updates to work properly.

We illustrate below the differing dynamics of SMS and SNEP when applied to a Bayesian logistic regression model with simulated data. We took a similar setup as in the SMS paper (Xu et al., 2014), generating a dataset $\mathcal{D} = \{(z_c, y_c)\}_{c=1}^{N}$ where covariates $z_c \in \mathbb{R}^d$ and response $y_c \in \{0, 1\}$. The conditional distribution of $y_c$ given $z_c$ and weights $x$ is

$$p(y_c = 1 \mid z_c, x) = \sigma(z_c^\top x) \tag{26}$$

where $\sigma$ denotes the logistic function. We used a Gaussian prior $p_0(x) = \mathcal{N}(x; 0_d, 10 I_d)$ on $x$ and the aim is to construct an approximation to the posterior $p(x \mid \mathcal{D})$. We generated $N = 50000$ data points with $d = 50$ using iid draws for the covariates, $z_c \sim \mathcal{N}(\mu, \Sigma)$ where $\mu \in [0,1]^d$ and $\Sigma = PP^\top$ with $P \in [-1,1]^{d \times d}$, with entries drawn uniformly at random for both $P$ and $\mu$. The generating weight vector $x^*$ is drawn from the prior $\mathcal{N}(x; 0_d, 10I_d)$. The labels $y_c$ are then sampled according to the model.

---

3. `http://julialang.org`.
4. `https://github.com/pluskid/Mocha.jl`.

Both algorithms were run with three workers each with one third of the data. SNEP is run with 1 inner loop iteration per synchronisation with the master (for the purpose of comparing against SMS). Varying numbers of MCMC samples per inner loop iteration were used for both algorithms, to investigate the effect of Monte Carlo noise on the performances of the algorithms (low number of samples meaning high noise and both lower performance and lower computational cost). The damping for SMS and the learning rate for SNEP were tuned for best performances. As the base exponential family, we used a full-covariance Gaussian. We compared the predictive RMSE $\sqrt{\sum_c |\hat{p}_c - y_c|^2/N}$ obtained with both methods over time where $\hat{p}_c = \sigma(z_c^T \bar{x})$ where $\bar{x}$ is the currently estimated posterior mean. We also compared the relative difference between the estimated posterior mean and that estimated from a long run of the No-U-Turn sampler (Hoffman and Gelman, 2014) as implemented in Stan (Carpenter et al., 2017) with 4 chains and 50000 iterations.

Figure 2 illustrates the performances of both algorithms and how they are influenced by the number of MCMC samples used per iteration. It can be observed that SNEP is more robust and better performing than SMS in the presence of noise. In general, we found that the number of samples needed for SMS to perform well grows with the dimensionality of the problem. In models with very high dimensionality and multi-modality (e.g., neural networks), the number of samples needed per step is very large leading to iterations that are computationally impractical, whereas SNEP can afford to use many fewer samples.

## 6.2 Bayesian Neural Networks

In this section we report experimental results applying SNEP and the posterior server to distributed Bayesian learning of neural networks. We have found that the SMS algorithm exhibited significant instabilities and was not suitable for these larger scale problems. Instead our aim here is to explore the behaviour of SNEP when varying various key hyperparameters. We used diagonal covariance Gaussians as the approximating exponential family due to computational cost considerations. While a diagonal Gaussian approximation can be quite poor for the full Bayesian posterior, past works have shown that they can produce good predictive performances (Hernandez-Lobato and Adams, 2015, Blundell et al., 2015, Kirkpatrick et al., 2017). Hence our aim is to evaluate the predictive performance of SNEP, treated as a distributed learning algorithm for neural networks, and comparing it to Adam (Kingma and Ba, 2015), a state-of-the-art stochastic gradient descent (SGD) algorithm with access to the whole data set on a single computer, as well as several state-of-the-art distributed SGD algorithms: asynchronous SGD (A-SGD) (Dean et al., 2012) and elastic averaging SGD (EASGD) (Zhang et al., 2015a).

In our experiments we used stochastic gradient Langevin dynamics with an preconditioning scheme reminiscent of Adam as the MCMC sampler. The preconditioning scheme is the same as the one proposed by (Li et al., 2016) except for the addition of a debiasing reminiscent of Adam. We found that sometimes this adaptive scheme lead to large injected noise, especially at the start of training which was detrimental to learning. To mitigate this effect we limited the standard deviation of the injected noise to be at most $\epsilon$ in our experiments. For further details see Appendix B.

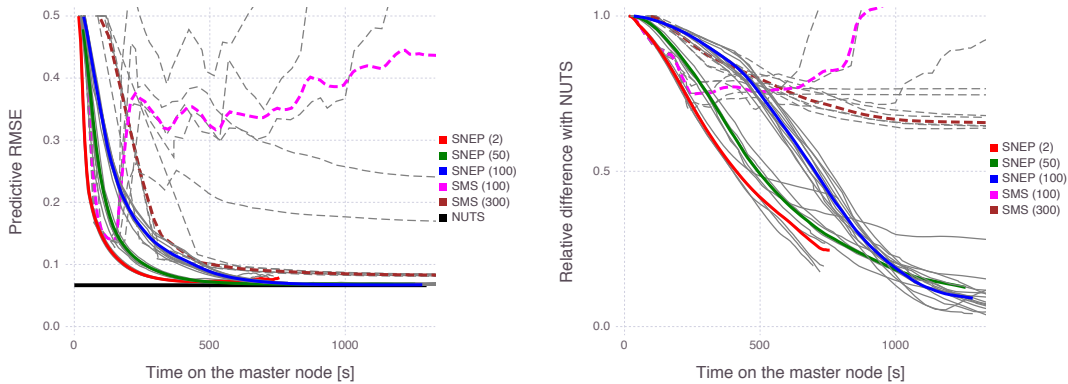

Figure 2: (left) Comparison of the predictive RMSE obtained with SMS (dashed-lines) and SNEP (full lines) versus the time on the master node when varying the number of MCMC samples per iteration (numbers reported in the legend). When too few samples are used for the moment estimates, SMS becomes unstable whilst SNEP does not suffer from this. SNEP is also consistently faster and more accurate than SMS. The accuracy obtained with the posterior mean estimated with a long run of Stan/NUTS is also displayed as comparison (horizontal line). (right) Relative difference between the posterior means estimated using Stan/NUTS and using SMS (dashed lines) or SNEP (full lines). Each coloured line corresponds to an average over several runs (represented in light grey).

### 6.2.1 MNIST, TWO LAYER FULLY CONNECTED NETWORK

In this section we look at training a fully connected neural network on the MNIST data set[5], which consists of 60000 training images of handwritten digits of size $28 \times 28$ and 10000 test images, the task being to classify each image into one of ten classes. The network has two hidden layers with 500 and 300 rectified linear units (ReLUs) respectively and softmax output units.

In the first set of experiments, we varied a number of hyperparameters of the learning regime while keeping the rest kept at default values, to investigate the sensitivity of the learning to these hyperparameters. The default values were chosen by hand in a rough initial round of experimentation as follows: the minibatch size is 100, the learning rate for SNEP was 0.02 and step size for the adaptive SGLD sampler was 0.001, the number of inner loop iterations per communication with master is $N_{sync} = 10$, the number of inner loop iterations per outer loop update is $N_{outer} = 10$, and the parameters were initialised with the popular Xavier initialisation (Glorot and Bengio, 2010). We set a minimum variance of 0.01 for the likelihood approximations, as we found that the estimated variances are otherwise too small due to the diagonal Gaussian approximation. Unless otherwise stated, we used four workers to explore the distributed behaviour. Test curves were produced by evaluating networks based on the approximate posterior means at the master. Each reported curve in the figures is an average over ten repetitions. In our experiments we used one CPU core per worker process. See Figure 3 for the results of this set of experiments.

---

5. `http://yann.lecun.com/exdb/mnist/`.

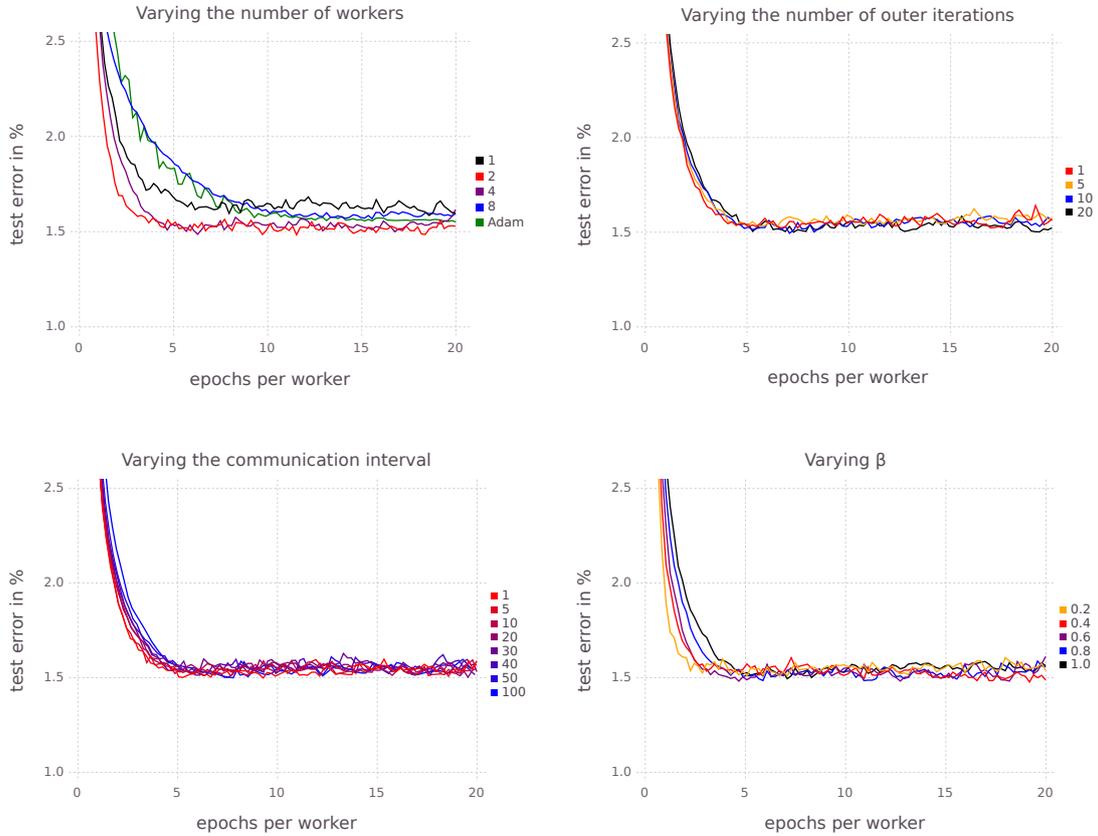

Figure 3: Results on a two-layer densely connected model on MNIST varying the parameters of SNEP and comparing against SGD, A-SGD, and EASGD. Four workers were used in the distributed methods in the experiments that held the number of workers fixed. This demonstrates that SNEP is insensitive to the full convergence of the outer loop, is robust to large communication intervals, that convergence is faster for a lower $\beta$ (on this data set), and that SNEP produces competitive results quickly. Detailed analyses of sub-figures (left to right, top to bottom) are reported in the main text.

First, we examined the behaviour of SNEP as $N_{worker}$ is varied. We see that two workers perform significantly better than one worker or Adam. However, performance deteriorates as the number of workers is increased further. This is possibly due to smaller number of data items per worker which makes the variational approximation more severe. Another factor we found is that because we imposed a minimum variance in for the likelihood approximations, using more workers gives tighter cavity distributions which can affect the scale of the natural-gradients and slow convergence significantly.

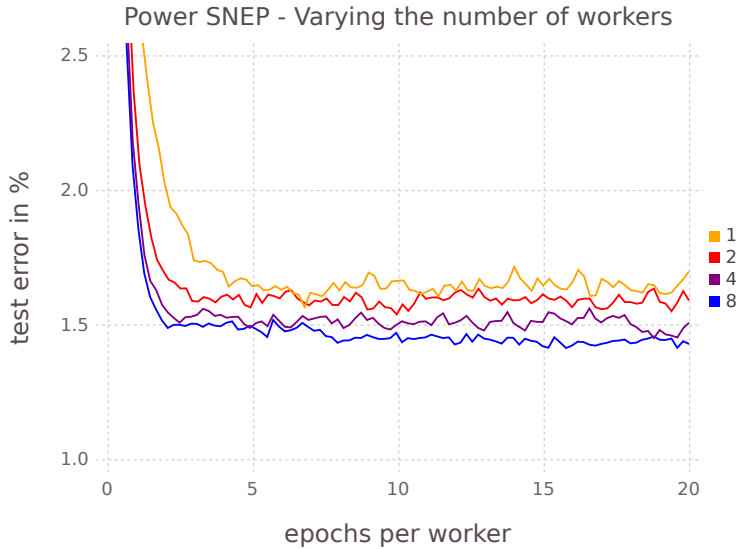

Figure 4: Results on a two-layer densely connected model on MNIST for power SNEP ($\beta = 1/N_{workers}$) varying the number of workers. There is a clear improvement in terms of convergence speed and final accuracy for more workers.

Second, we investigated the effect of varying $N_{sync}$ and $N_{outer}$. Figure 3 demonstrates that SNEP is insensitive to these hyperparameters over reasonably large range of values. Note in particular that infrequent communications with the posterior server (up to once every 100 iterations in this experiment) did not significantly deteriorate the learning process at all. In fact, the related SMS algorithm (Xu et al., 2014) effectively involves communications with the master once every thousands of iterations.

Third, we investigated the effect of varying the $\beta$ parameter. Interestingly, lower values of $\beta$ sped up convergence significantly. The literature on power EP links the $\beta$ parameter with locally minimising $\alpha$−divergences (where $\beta = 1/\alpha$). Lower values of $\beta$ correspond to local approximations that are likely to capture more of the full posterior rather than matching one mode (Minka, 2005). Another factor is that given the minimum variance setting, lower values of $\beta$ increase the variance of the cavity distribution allowing for more local exploration. Generally, In our experiments we found that a heuristic of setting $\beta = 1/N_{workers}$ worked particularly well for MNIST. We call the resulting algorithm power SNEP (p-SNEP). With this parameter setting we see a clear advantage for more workers (see figure 4). More workers give better results in terms of the test error and convergence is fastest for eight workers. Note that eight workers reach a test error of 1.5% after about two epochs, which is much faster than our experiments with $\beta = 1.0$.

Finally, we compared SNEP with two distributed SGD algorithms, asynchronous SGD (A-SGD, also known as Downpour) (Dean et al., 2012) and elastic averaging SGD (EASGD) (Zhang et al., 2015a). Our implementations differ slightly from those given in the papers. For A-SGD, both the workers and master performed Adam updates with independently chosen step size constants. The authors of EASGD do not address the case of distributing the data among the workers. Nonetheless, we found this posed no problem to the algorithm. All hyperparameters were tuned to optimise performances. We used the same length of communication intervals $N_{sync} = 10$ and four workers for all algorithms. As shown in figure 5, SNEP (with $\beta = 1$) and p-SNEP (with $\beta = 1/N_{workers}$) achieve performances comparable to these state-of-the-art methods, both in terms of speed and final test accuracies. SNEP is comparable to Adam and better than A-SGD. p-SNEP outperforms A-SGD, Adam and SNEP, and is slightly faster initially than EASGD but achieves slightly higher test errors. It is important to note that we are tackling a harder problem with SNEP/p-SNEP, that of Bayesian learning rather than optimisation.

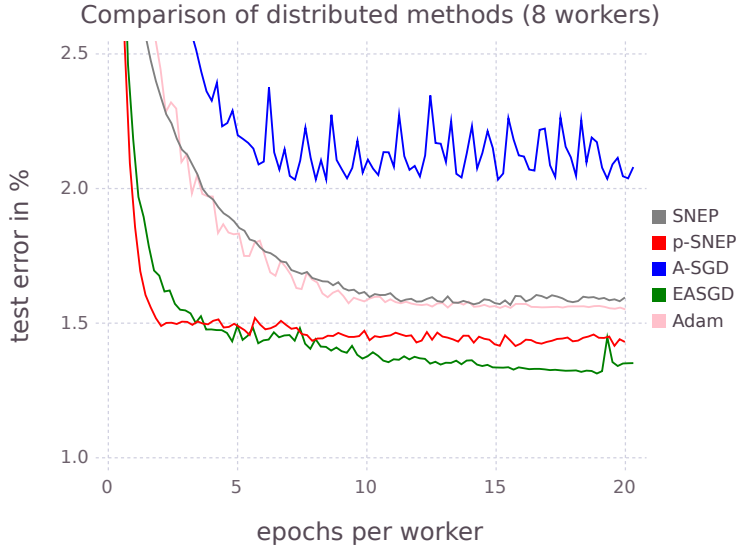

Figure 5: Results on a two-layer densely connected model on MNIST comparing SNEP and p-SNEP to asychronous SGD and elastic averaging SGD.

### 6.2.2 MNIST, Very Deep Fully Connected Network

Deeper models pose a much harder learning problem—as the depth increases, the number of saddle points increases exponentially (Dauphin et al., 2014). Moreover, learning often gets stuck not at critical points but in large plateaus of the error surface (Lipton, 2016). It is, therefore, interesting to evaluate (p-)SNEP against standard methods on a deeper fully-connected network, and we chose the architecture from Neelakantan et al. (2016), which has 20 hidden layers of 50 units each.

The experimental setup was similar to the previous section. The distributed methods used a communication intervals of $N_{sync} = 10$ and a varying number of workers. For p-SNEP, we optimized the prior variance, scaling of the injected noise, learning rate, step size, and beta hyperparameters by a course grid search over five orders of magnitude.

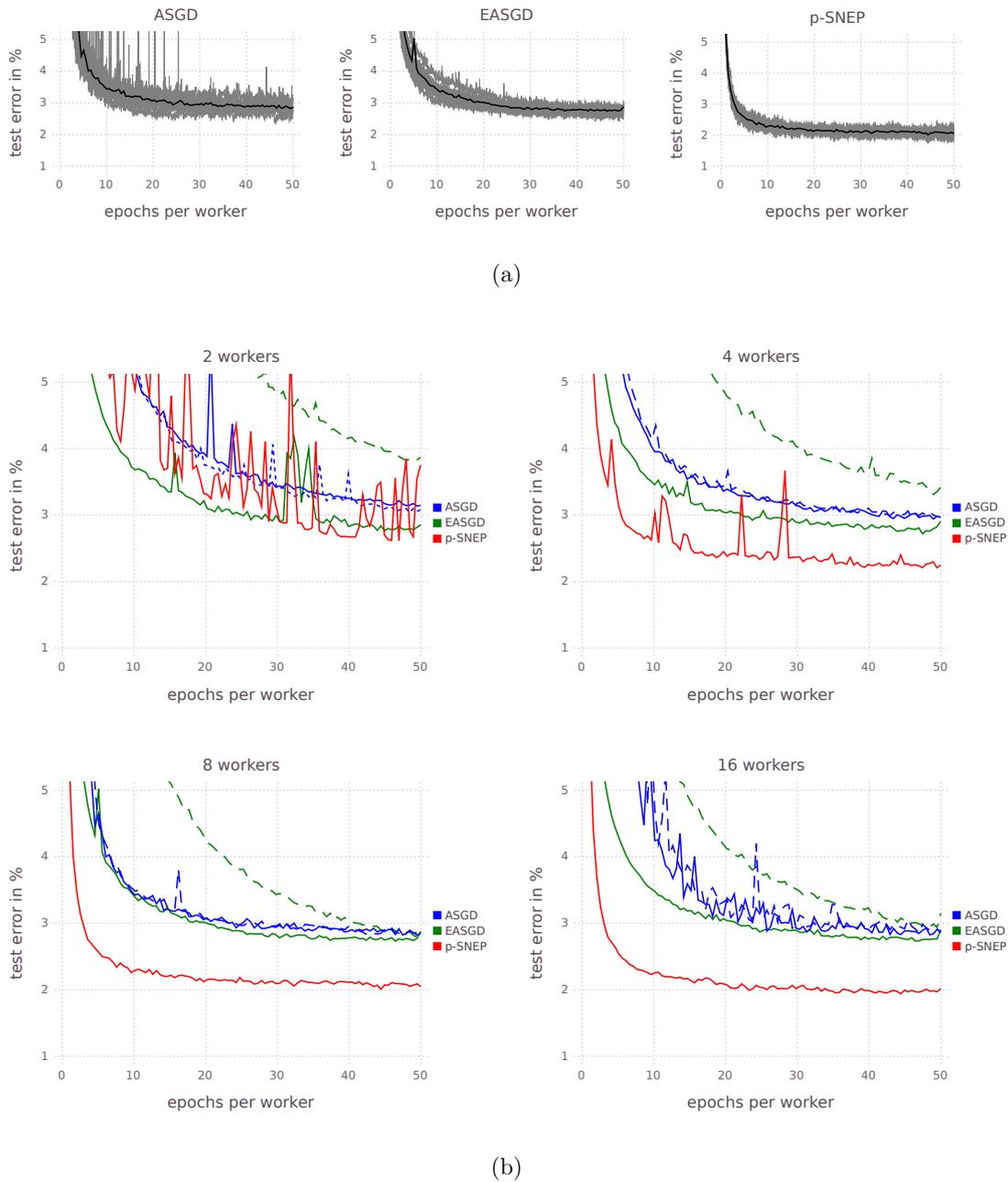

(a)

(b)

Figure 6: Results on deep narrow model for MNIST. (a) Comparing distributed methods for 8 workers. p-SNEP attains an extra percent of accuracy with less variability in the runs. The solid line is the average over 10 runs. (b) Varying the number of workers, averaged over 10 runs. The dashed line indicates result with no prelearning phase for A-SGD and EASGD, whereas for those methods the solid line indicates an Adam pre-learning phase of a third of an epoch.

Similarly, for A-SGD we optimized the Adam step size constants, separately on the workers and master, for EASGD the moving rate and Adam step size constants, and for SGD the Adam step size constants and scaling of injected noise, from no noise to three orders of magnitude. All methods were initialized by the Xavier method, and run with and without a non-distributed pre-learning phase of a third of an epoch. We found that pre-learning was essential for EASGD, while it was unnecessary for A-SGD and p-SNEP. In fact, for SNEP, too long a pre-learning phase is actually harmful to the final solution reached, which we suggest is due to the reduced exploration between workers. We only report results for p-SNEP without pre-learning (as per the previous section).

The same Adam step size constant was found to be optimal across all methods, 0.00065. For p-SNEP, the prior variance was set to 0.05, $\beta$ to 1/8, and the learning rate to 0.04. For EASGD, the moving rate was set to 0.25. For A-SGD, the worker step sizes were set to 0.00065 and the master step size to a third of this; it must be slower to account for the staleness of the parameters. For SGD, we discovered that a small amount of noise, the same constant as is optimal for p-SNEP, reduces the variance of the runs around their mean by helping some runs avoid getting stuck in bad solutions. However, SGD did not converge to within the plot limits within the time set for experiments and so is omitted in the results shown in Figure 6a.

p-SNEP was found to significantly outperform A-SGD and EASGD, obtaining an extra percent of accuracy relative to other methods after fifty epochs for 8 and 16 workers. From the perspective of traditional neural network learning, our algorithm has two advantages for learning deep models: the addition of noise, and a principled method of regularizing the parameters. As in Neelakantan et al. (2016), we found it was important to add noise to the SGD methods. Although with a good initialization, added noise only improved the worst runs and not the best, smoothing out convergence. We conjecture that the added noise helps learning escape the difficult saddle points and plateaus. For this model, a strong prior variance was found to be important for p-SNEP to extract the full capacity of the model. We conjecture that this allows the gradients to flow back through the network more easily by forcing the parameters of the top layers to be small. Surprisingly, we found L2 regularization did not help with A-SGD and EASGD. p-SNEP also clearly benefits from additional workers in this setting, in contrast to A-SGD and EASGD (see Figure 6b).

It is interesting to note that this model has about one sixth the parameters of the previous two layer model. Yet, due to the deeper architecture and ability to successfully navigate the difficult learning landscape, p-SNEP is able to obtain similar test accuracies.

### 6.2.3 CIFAR-10, Convolutional Network

We also experimented with distributed Bayesian learning of convolutional neural networks, applying these to the CIFAR-10 data set (Krizhevsky, 2009) which consists of 50000 training instances and 10000 test instances from 10 classes, each instance being a 32x32 colour natural image. The network used is the one described in Alex Krizhesky's CIFAR tutorial[6] and consists of 8 layers: a first convolutional layer followed by max-pooling and local response normalization, a second convolutional layer also followed by max-pooling and local response normalization, and a third convolutional layer followed by a fully connected layer.

---

6. `https://code.google.com/p/cuda-convnet/wiki/Methodology`

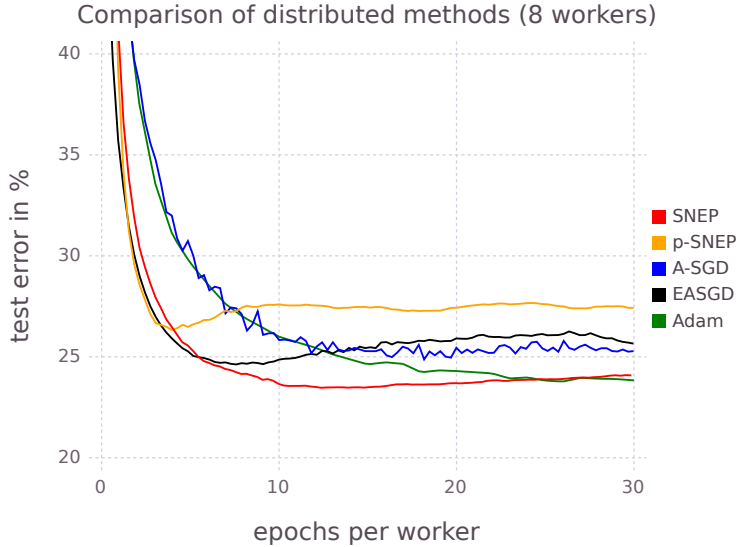

Figure 7: Learning curves varying the number of workers (averaged over three runs). The green line shows Adam with default parameters for comparison. Best viewed in colour.

For SNEP, we used the following settings: $N_{sync} = 10$, $N_{outer} = 10$, mini-batch size 100, weights initialised with the Xavier initialisation, and SNEP learning rate and adaptive SGLD step sizes of 0.02. We did not investigate improving performance by building in invariances using data perturbations or having multiple learning phases with different learning rates. Learning curves averaged over 3 runs are shown in Figure 7. All distributed algorithms are shown with hand-tuned optimal hyperparameter settings. On this data set SNEP outperforms other distributed algorithms. It converges as quickly as EASGD and p-SNEP but reaches a better final test accuracy. p-SNEP converges very fast but to a worse local optimum. It is not clear why p-SNEP performed better on previous experiments but not here; understanding the optimal choice of $\beta$ is an interesting area for future research as we have observed that this parameter has a strong influence on the performance of the algorithm.

## 7. Discussion

In this paper, we have proposed a novel alternative to expectation propagation called stochastic natural-gradient expectation propagation (SNEP). SNEP is demonstrably convergent, even when using Monte Carlo estimates of the moments/mean parameters of tilted distributions. Experimentally, we find that SNEP converges more efficiently and more stably than other methods considered, particularly when Monte Carlo noise is high. Using SNEP, we have proposed the posterior server architecture for distributed Bayesian learning using an asynchronous non-blocking message-passing protocol. The architecture uses a separate MCMC sampler on each worker, and SNEP to coordinate the samplers across the cluster so that the target distributions agree on the moments which characterise the base exponential family. In contrast with typical maximum likelihood parameter server architectures, the posterior server allows each worker to learn separate variational parameters, and as a result requires less frequent synchronisation across the cluster. We believe that this insight can allow for significant advances to distributed learning, although more work is still needed to make this reality. Finally, we applied SNEP and the posterior server to distributed Bayesian learning of both fully-connected and convolutional neural networks, where we showed performances on par with or better than state-of-the-art distributed and non-distributed optimisation algorithms. In conclusion, we have demonstrated that Bayesian learning of large complex models can be achieved efficiently and effectively in a distributing computing setting, and can achieve state-of-the-art performances on learning neural networks.

Our work leaves open a number of interesting avenues of future research, which we have discussed throughout the paper. It would be interesting to develop other novel convergent alternatives to EP with potentially faster convergence (e.g. using a single loop instead of double loop algorithm), and to apply such methods to other settings where Monte Carlo estimates are used within EP. It would also be interesting to further investigate combining SNEP with ideas from other projects such as the parameter tying idea of stochastic and averaged EP. We also believe that further explorations of learning regimes and hyperparameters, as well as of larger data sets, is called for, and will demonstrate further improvements to the method. These explorations can include using samples obtained at workers to form predictive probabilities, evaluating the algorithms using test log probabilities and on their abilities to quantify uncertainties, understanding the role of the $\beta$ parameter, and combining SNEP with simulated annealing for optimisation. Another interesting area for application is on-device machine learning which falls naturally into our distributed data set-up. It would be interesting to compare SNEP to existing approaches such as federated learning (McMahan et al., 2016) We have also noted that our application of SNEP to neural network learning necessitates a diagonal covariance Gaussian approximation for computational reasons. Further work can consider richer posterior approximations, such as block-diagonal covariances or the use of matrix variate Gaussian distributions (Louizos and Welling, 2016). Ultimately, we believe these explorations will demonstrate the utility of a distributed Bayesian approach to learning.

## Acknowledgments

LH is funded by the EPSRC OxWaSP CDT through grant EP/L016710/1. SW gratefully acknowledges support from the EPSRC AIMS CDT through grant EP/L015987/2. TL gratefully acknowledges funding from the Scatcherd European scholarship and EPSRC Grant EP/L505031/1. SJV thanks EPSRC for funding through EPSRC Grants EP/N000188/1 and EP/K009850/1. YWT gratefully acknowledges EPSRC for research funding through grant EP/K009362/1, and the European Research Council under the European Union's Seventh Framework Programme (FP7/2007-2013) ERC grant agreement no.

617071. The authors thank Daniel Hernández-Lobato, Yingzhen Li, Thang Bui and Rich Turner for valuable feedback on the preprint and suggestions about tied EP factors.

## Appendix A. Relationship of Ideal Variational Problem in the Extended Exponential Family to Variational Inference

As expected, the ideal variational problem in the extended exponential family in (3) is intractable and approximations are needed for tractability, with different approximations leading to different variational approaches. For readers unfamiliar with this framework, it might be illuminating to link the above framework to standard mean field variational inference, which corresponds to approximating the mean domain $\tilde{\mathcal{M}}$ by the set of moments achievable by some family of factorised distributions $q$ over $x$. The standard evidence lower bound on the log marginal probability (also known as the variational free energy) is,

$$
\begin{aligned}
p(\{D_i\}_{i=1}^n) &\geq \mathbb{E}_q[\log p(x, \{D_i\}_{i=1}^n) - \log q(x)] \\
&= \mathbb{E}_q\left[\theta_0^\top s(x) - A(\theta_0) + \sum_{i=1}^n \ell_i(x)\right] - \mathbb{E}_q\left[\log q(x)\right] \\
&= \tilde{\theta}^\top \tilde{\mu} - A(\theta_0) - \mathbb{E}_q\left[\log q(x)\right]
\end{aligned}
$$

where we have introduced $\mu = \mathbb{E}_q[s(x)]$ and $\nu_i = \mathbb{E}_q[\ell_i(x)]$ as the moments of $q$, and we have used the definition of the extended natural and mean parameters $\tilde{\theta}$, $\tilde{\mu}$. The second term is a constant not dependent on $q$, while the third term above is the negative entropy, so that the above is equivalent to (3) except that the optimisation is only over a family of factorised distributions $q$.

## Appendix B. Additional Techniques for Bayesian Neural Networks

In our experiments we investigated the use of SNEP and the posterior server for distributed Bayesian learning of neural networks. To get the method working well on the notoriously complicated posterior distributions for neural networks, a number of additional techniques are needed, which we describe here.

### B.1 Adaptive Stochastic Gradient Langevin Dynamics

Most of the computational costs associated with the algorithm involve the MCMC updates to the state $x_i$. When the number of data points stored on each compute node is large, standard MCMC updates are infeasible as each update requires computations involving every data point. In our experiments we used the stochastic gradient Langevin dynamics (SGLD) algorithm proposed by Welling and Teh (2011) which scales well to large data sets.

SGLD uses mini-batches of data to provide unbiased estimates of gradients which are used in a time discretized Langevin dynamics simulation whose stationary distribution is the desired tilted distribution (20). SGLD injects noise in every step. As the discretization stepsize decreases the noise due to the stochastic gradients is eventually dominated by the injected noise and hence neglibible. Likewise the discretization introduces errors which tend to zero as the discretization step sizes decreases to zero; see (Teh et al., 2015, Vollmer et al., 2016). Recall that the data points on compute node $i$ is $D_i = \{y_c\}_{c \in S_i}$, and the log likelihood is

$$\ell_i(x_i) = \sum_{c \in S_i} \log p(y_c \mid x_i).$$

Let $B^{(t)} \subset S_i$ be a mini-batch of data, chosen uniformly at random with fixed size. Each SGLD update is,

$$x_i^{(t+1)} = x_i^{(t)} + \kappa_t (M^{(t)})^{-1} \left( \nabla s(x_i^{(t)})^\top (\theta_i' - \beta_i^{-1} \lambda_i^{(t)}) + \beta_i^{-1} \frac{|S_i|}{|B^{(t)}|} \sum_{c \in B^{(t)}} \nabla \log p(y_c | x_i^{(t)}) \right) + \eta_t,$$

$$\eta_t \sim \mathcal{N}(0, 2\kappa_t (M^{(t)})^{-1}). \tag{27}$$

The term inside the parentheses is an unbiased estimate of the gradient of the log density of the tilted distribution (20), $\kappa_t$ is the discretization step size, $M^{(t)}$ is (an adaptive) diagonal mass matrix, while $\eta_t$ is an injected normally-distributed noise, which prevents SGLD from converging to a mode of the distribution and distinguishes it from stochastic gradient descent. See Welling and Teh (2011) for details.

In (27), $M^{(t)}$ is a mass matrix which effectively controls the length scale of updates to each dimension of $x_i$. It is well known that in neural networks the length scales of gradients differ significantly across different parameters and adaptation of learning rates specific to each parameter is crucial for successful deployment of stochastic gradient descent learning. We have found that this is the case for SGLD as well, and used an adaptation scheme for the mass matrix reminiscent of Adam (Kingma and Ba, 2015). The adaptation scheme is the same as the one proposed by (Li et al., 2016) except for the addition of a debiasing reminiscent of Adam. At iteration $t$ let $g_t$ be the stochastic gradient estimate in (27). We use a diagonal mass matrix $M_t$ that is updated according to the following update equations:

$$v_t = \beta v_{t-1} + (1 - \beta) g_t \odot g_t$$

$$\text{diag}(M_t) = \frac{v_t}{1 - \beta^2},$$

where $\odot$ denotes the elementwise product. We used $\beta = 0.999$ in our experiments. The actual adaptive stepsize is then given by

$$\kappa_t = \frac{\epsilon}{\sqrt{M_t} + \delta},$$

where $\epsilon = 10^{-3}$ and $\delta = 10^{-8}$ is added for numerical stability. We found that sometimes this adaptive scheme lead to large injected noise, specially at the start of training which was detrimental to learning. To mitigate this effect we limited the standard deviation of injected noise to be at most $\epsilon$ in our experiments. An alternative solution to this problem could be the approach proposed by Lu et al. (2017).

## B.2 Shifting MCMC States After Communication with Posterior Server

After each communication with the posterior server, the target distribution of the MCMC sampler on the worker, say $i$, is changed, because of $\theta_{-i}$ being updated in Step 16 of

Algorithm 1. Assuming that the MCMC sampler had previously converged, it will now not be so anymore because of this shift in the target distribution, and a number of burn-in iterations may be needed before the mean parameter estimates can be used for SNEP updates again.

For Gaussian base exponential families, we can shift the MCMC state along with the target distribution when $\theta_{-i}$ is updated in the following way. Suppose $\mu_i^{\text{old}}$, $\Sigma_i^{\text{old}}$, $\mu_i^{\text{new}}$, $\Sigma_i^{\text{new}}$ are the means and covariances of the approximate Gaussian posterior before and after the update to $\theta_{-i}$ (with natural parameters $\lambda_i + \theta_{-i}^{\text{old}}$, $\lambda_i + \theta_{-i}^{\text{new}}$ respectively where $\lambda_i$ is the current natural parameter of the Gaussian likelihood approximation). Suppose $x_i^{\text{old}}$ is the MCMC state before the update. Then we shift the MCMC state as follows:

$$x_i^{\text{new}} = \mu_i^{\text{new}} + (\Sigma_i^{\text{new}})^{\frac{1}{2}} (\Sigma_i^{\text{old}})^{-\frac{1}{2}} (x_i^{\text{old}} - \mu_i^{\text{old}}). \tag{28}$$

The idea is that $x_i^{\text{new}}$ should be at the same location relative to the new Gaussian approximation to the posterior as $x_i^{\text{old}}$ is relative to the old Gaussian approximation. We have found that no burn-in is needed with this shift in the MCMC state after each communication.